\documentclass[10pt,twocolumn,letterpaper]{article}

\usepackage{cvpr}              % To produce the CAMERA-READY version
\usepackage[utf8]{inputenc} % allow utf-8 input
\usepackage[T1]{fontenc}    % use 8-bit T1 fonts
\usepackage{url}            % simple URL typesetting
\usepackage{booktabs}       % professional-quality tables
\usepackage{amsfonts}       % blackboard math symbols
\usepackage{nicefrac}       % compact symbols for 1/2, etc.
\usepackage{microtype}      % microtypography
\usepackage{xcolor}         % colors
\usepackage{wrapfig}
\usepackage{graphicx}
\usepackage[table]{xcolor}
\usepackage{amsmath,amssymb}
\usepackage{mathtools}
\usepackage{amsthm}
\usepackage{caption}
\usepackage{listings}
\usepackage{multirow}
\usepackage{subcaption}

\definecolor{cvprblue}{rgb}{0.21,0.49,0.74}
\usepackage[pagebackref,breaklinks,colorlinks,allcolors=cvprblue]{hyperref}

\def\paperID{} % *** Enter the Paper ID here
\def\confName{CVPR}
\def\confYear{2026}

\title{CLIP-Free, Label Free, Unsupervised Concept Bottleneck Models}

\author{
Fawaz Sammani$^{1,2}$, Jonas Fischer$^{1}$, Nikos Deligiannis$^{2}$ \\
$^{1}$ Max-Planck-Institut f{\"u}r Informatik, Saarbr{\"u}cken, Germany \\
$^{2}$ ETRO Department, Vrije Universiteit Brussel, Belgium \\
{\tt\small fawaz.sammani@vub.be, jonas.fischer@mpi-inf.mpg.de, ndeligia@etrovub.be}
}

\begin{document}
\maketitle

\begin{abstract}
Concept Bottleneck Models (CBMs) map dense feature representations into human-interpretable concepts which are then combined linearly to make a prediction. However, modern CBMs rely on the CLIP model to obtain image-concept annotations, and it remains unclear how to design CBMs without the CLIP bottleneck. Methods that do not use CLIP instead require manual, labor intensive annotation to associate feature representations with concepts. Furthermore, all CBMs necessitate training a linear classifier to map the extracted concepts to class labels. In this work, we lift all three limitations simultaneously by proposing a method that converts any frozen visual classifier into a CBM without requiring image-concept labels (label-free), without relying on the CLIP model (CLIP-free), and by deriving the linear classifier in an unsupervised manner. Our method is formulated by aligning the original classifier’s distribution (over discrete class indices) with its corresponding vision-language counterpart distribution derived from textual class names, while preserving the classifier’s performance. The approach requires no ground-truth image–class annotations, and is highly data-efficient and preserves the classifiers reasoning process. Applied and tested on over 40 visual classifiers, our resulting unsupervised, label-free and CLIP-free CBM (\textbf{U-F$^2$-CBM}) sets a new state of the art, surpassing even supervised CLIP-based CBMs. We also show that our method can be used for zero-shot image captioning, outperforming existing methods based on CLIP, and achieving state-of-art.  
\end{abstract}

\section{Introduction}
\label{intro}

Visual classifiers predict a class as a linear combination of dense, high‑dimensional visual feature vectors that are difficult to interpret by humans. Concept Bottleneck Models (CBMs) \citep{pmlr-v119-koh20a} address this challenge by mapping these feature vectors into a set of human-interpretable concepts, each associated with an activation score (referred to as a concept activation). Predictions are then made as a linear combination of these concept activations. Initial CBMs required image–concept annotations to train the bottleneck layer that maps dense features to concepts. Modern CBMs \citep{oikarinen2023labelfree, yuksekgonul2023posthoc, Benou2025ShowAT} overcome this limitation by leveraging the CLIP model \citep{Radford2021LearningTV} or vision–language grounding models \cite{srivastava2024vlgcbm} to provide image-concept annotations. Since CLIP models map image and text into a shared embedding space, these approaches can query images against an entire pool of predefined textual concepts and use cosine similarity scores to find a matching annotation. These approaches are commonly referred to as \textit{label-free CBMs}.

\noindent In many real-world settings, a high-performing, task-specific \textit{legacy} model already exists and often achieves strong results on the target task \citep{zhang2024why}. When CLIP is used to generate image-concept annotations, the resulting CBM is anchored to CLIP’s embedding space; the legacy model must be interpreted through CLIP’s notion of similarity, rather than through its own learned representation. In this sense, the approach is not fundamentally different from replacing the legacy model with a CLIP-based CBM. Furthermore, this dependence can transfer CLIP’s biases and behavior into the legacy model (e.g., importing the CLIP typographic bias \cite{goh2021multimodal} into a DINO model). A natural question that then arises is: \textit{how to develop a label-free CBM for such legacy specialist models without the CLIP constraint?} Retraining such a specialist on a large image–text corpus following the CLIP approach is impractical, in terms of computational cost and need of a huge amount of image-text data. Furthermore, obtaining ground-truth image-concept annotations through manual human-labor is time consuming and expensive. Finally, retraining this legacy model further alters its original decision-making process and distribution, which is typically not desired.

\noindent In this work, we lift these limitations by first proposing a method dubbed as TextUnlock, which aligns a distribution of a frozen visual classifier to its corresponding vision-language counterpart, without relying on CLIP. TextUnlock has four important properties: First, it is \textit{efficient}; it is inexpensive to train and can be performed on any standard hardware, regardless of the size of the original classifier. The number of data points is also significantly reduced compared to CLIP-based approaches. Secondly, it is \textit{label-free}, no labels are required to achieve this formulation. Thirdly, TextUnlock is \textit{trained to preserve} the original distribution and reasoning process of the classifier, and does not compromise the classifier's original performance (average of 0.2 points drop in accuracy). Finally, our method is applicable to \textit{any} vision architecture, whether convolutional-based, transformer-based or hybrid. After the original classifier's distribution is aligned to its vision-language counterpart distribution with TextUnlock, we simply query the transformed classifier's image features against a set of predefined text concepts to obtain concept activations for our CBM, and then derive the concept-to-class classifier directly from the transformed classifier’s text source, without requiring any additional training and operating in an unsupervised manner (i.e., we do not train a linear probe to associate the concept activations to class labels).

\begin{figure*}[ht]
    \centering
    \includegraphics[width=\textwidth]{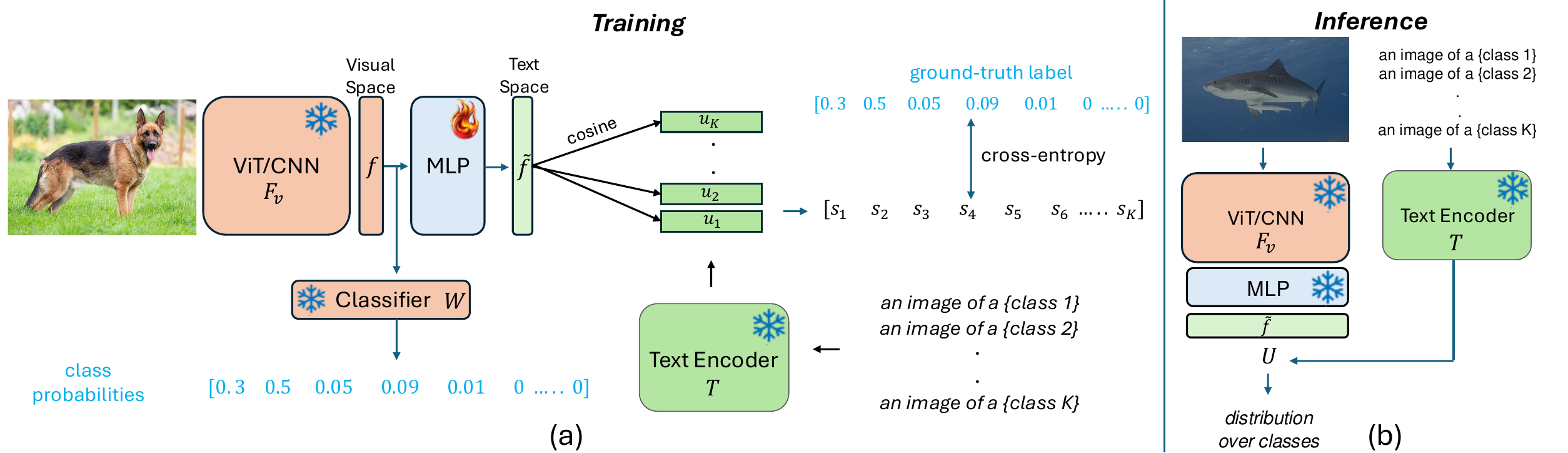}
    \caption{\textbf{Overview of our proposed TextUnlock.} \textbf{(a)} The process of training the MLP mapping between vision and text space. \textbf{(b)} The process of inference with the adapted visual classifier. The text encoder acts as weight generator for  a linear classifier. \includegraphics[height=1.5ex]{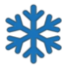} indicates that the module is frozen, while \includegraphics[height=1.5ex]{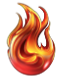} indicates trainable. }
    \label{method}
    %\vspace{-0.5cm}
\end{figure*}

In summary, our contributions are as follows: \textbf{(i)} We propose a method to convert any frozen visual classifier into a CBM without relying on the CLIP constraint (CLIP-free), neither on annotated image-concept data (label-free), and for the first time, also deriving the concept-to-class classifier in a fully unsupervised manner. \textbf{(ii)} We demonstrate the effectiveness of our method with 40 different architectures, along with extensive ablation studies and intervention results, and further show how our method can also be used to perform zero-shot image captioning. \textbf{(iii)} We set new state-of-the-art results on CBMs, outperforming existing works including supervised CLIP-based CBMs, despite being trained only on ImageNet-1K.

\section{Related Work}
Concept Bottleneck Models (CBMs) are inherently interpretable models that predict by first mapping inputs to human-understandable concepts and then combining those concepts linearly to make the final decision, originally requiring concept annotations for each image \cite{oikarinen2023labelfree}. Recently, Label-Free CBMs (LF-CBMs)  have been introduced, which leverage CLIP to provide image-concept annotations. CLIP is used to generate ground-truth image–concept similarity scores, which serve as supervision for training the bottleneck layer of a downstream classifier. Several further works \citep{, yuksekgonul2023posthoc, Benou2025ShowAT} adopted this pipeline. Another work uses vision-language grounding models \cite{srivastava2024vlgcbm} to obtain image-concept annotations. Another line of work \citep{Yang2022LanguageIA, Panousis2023SparseLC, ao2024DiscoverthenNameTC, Knab2024DCBMDV} builds CBMs directly on top of CLIP, making them specifically tailored to CLIP-only models and thereby eliminating the need for a separate bottleneck training stage. In these methods, image features are queried against a predefined set of concepts in the CLIP embedding space, and the resulting cosine similarities are used directly as concept activations. As a result, all the above methods either (1) are limited to CLIP models, or (2) depend on CLIP or vision–language models to obtain the image-concept annotations. On the other hand, we propose a fully CLIP-free CBM; our method does not rely on any external vision-language model to generate the image-concept annotations, nor on human annotation, and can be applied to any legacy model without the CLIP constraint. Finally, all CBMs works to date require training a linear probe to map concept activations to class predictions. In contrast, we demonstrate that our method can be readily used to derive the linear probe in a fully unsupervised manner.

\noindent \textbf{Transforming Visual Features to Text:} There exist works that try to decode visual features of models beyond CLIP. DeVIL \citep{Dani2023DeViLDV} and LIMBER \citep{merullo2023linearly} train an autoregressive text generator to map visual features into image captions, leveraging annotated image-caption pairs as ground-truth data. Notably, all these works (1) rely on annotated datasets and (2) explicitly train the generated text to align with what annotators want the visual features to describe, thereby altering the classifier’s original reasoning process. Text-to-Concept (T2C) \citep{Moayeri2023TextToConceptB} also trains a linear layer to map image features of any classifier into the CLIP vision encoder space, such that they can be interpreted via text using the CLIP text encoder. Notably, all these methods rely on the CLIP approach and/or its supervision, and they alter the classifier’s distribution by entirely discarding its output class distribution. In contrast, our method is CLIP-free, can be applied to any pretrained classifier without requiring any annotated data, and explicitly preserves the classifier’s predictive distribution.

\section{Method}
\label{sec:method}

We first elaborate on our proposed TextUnlock method in Section \ref{subsec:textunlock}, which will then allow us to design our proposed U-F$^2$-CBM, which we discuss in Section \ref{sec:zscbms}. \textbf{U} stands for unsupervised; \textbf{L$^2$} for CLIP-free and Label-free (hence ‘double free’).

\subsection{TextUnlock}
\label{subsec:textunlock}
A visual classifier assigns an image to a specific category from a predefined set of discrete class labels. For example, in ImageNet-1K \citep{Deng2009ImageNetAL}, this set contains $1,000$ class labels. Originally, these discrete class labels correspond to class names in text format. For example, in ImageNet-trained models, the discrete label $1$ corresponds to the class \textit{goldfish}. These classes are typically discretized to facilitate training with cross-entropy. However, when the textual class names are used, they provide an advantage. Specifically, when the textual class names are embedded into vector representations (e.g., using a word embedding model or text encoder), they provide semantic information. Specifically, These embeddings reside in a continuous space where nearby vectors capture related semantic associations. For the ``goldfish” example, neighboring vectors might include terms such as ``freshwater”, ``fins” and ``orange”. Such associations can be viewed as high-level conceptual attributes that characterize the class. Our method learns to map images into this text embedding space using \emph{only} class names, thus linking both the class name \emph{and its surrounding semantic associations} with the image. This process thus naturally supports unseen words that are not part of the class names (e.g., concepts in the CBM). We accomplish this through a trainable multilayer perceptron (MLP) that projects the visual features into the text embedding space, and is explicitly trained to match its distribution with the original classifier’s class distribution. This is done while keeping both the visual and textual encoders frozen. By using solely the class names without any supplementary information, we can learn a semantically meaningful image-text space. This allows us to query the visual classifier with text queries beyond the class names (e.g., concepts) and obtain concept activations for CBMs. 

\noindent Consider an image $I$ and a visual classifier $F$ composed of a visual feature extractor $F_v$ and a linear classifier $W$. Note that $F$ can be of any architecture. $F_v$ embeds $I$ into an $n-$dimensional feature vector $f \in \mathbb{R}^{n}$. That is, $f = F_v(I)$. The linear classifier $W \in \mathbb{R}^{n \times K}$ takes $f$ as input and outputs a probability distribution $o$ for the image across $K$ classes. That is, $o = \text{softmax}(f.W) \in \mathbb{R}^{K}$. For ImageNet-1K, $K=1000$. Consider also any off-the-shelf text embedding model $T$ which takes in an input text $l$ and embeds it into a $m-$dimensional vector representation $u \in \mathbb{R}^{m}$. That is, $u = T(l)$. Note that $u$ and $f$ are not in the same space and can have a different number of dimensions, so we cannot query $f$ with the text $l$. 

\noindent We propose to learn a lightweight $\text{MLP}$ mapping function that projects the visual features $f$ into the text embedding space of $T$, resulting in a new vector $\tilde{f}$. That is, $\tilde{f} = \text{MLP}(f)$, where $\tilde{f} \in \mathbb{R}^{m}$. Note that the visual encoder $F_v$, the linear classifier $W$, and the text encoder $T$ are all frozen; only the \text{MLP} is trainable, making our method \textit{parameter-efficient}. We then take the textual class names of the $K$ classes, and convert each into a text prompt $l^p$, represented as: ``an image of a \{class\}" where \{class\} represents the class name in text format. This results in $K$ textual prompts, each of which is encoded with $T$: $u_i = T(l^p_i)$,  $\forall i = 1, \dots, K$. Stacking all the encoded prompts, we get a matrix $U \in \mathbb{R}^{K \times m}$. Here, $U$ acts as weights of the classification layer for our approach. We then calculate the cosine similarity\footnote{in the rest of this paper, we will omit the unit norm in cosine similarity to reduce clutter, and represent it with the dot product.} between each $u_i$ and the visual features $f$: $s_i = \tilde{f}.u_{i}$. Equivalently, this can be performed as a single vector-matrix multiplication: $S = \tilde{f}.U^T$, where $S \in \mathbb{R}^{K}$ represents the cosine similarity scores between the visual features and every text prompt $l^p_i$ representing a class. In other words, $S$ represents the classification logits of our approach.

\begin{figure*}[ht]
    \centering
    \includegraphics[width=\textwidth]{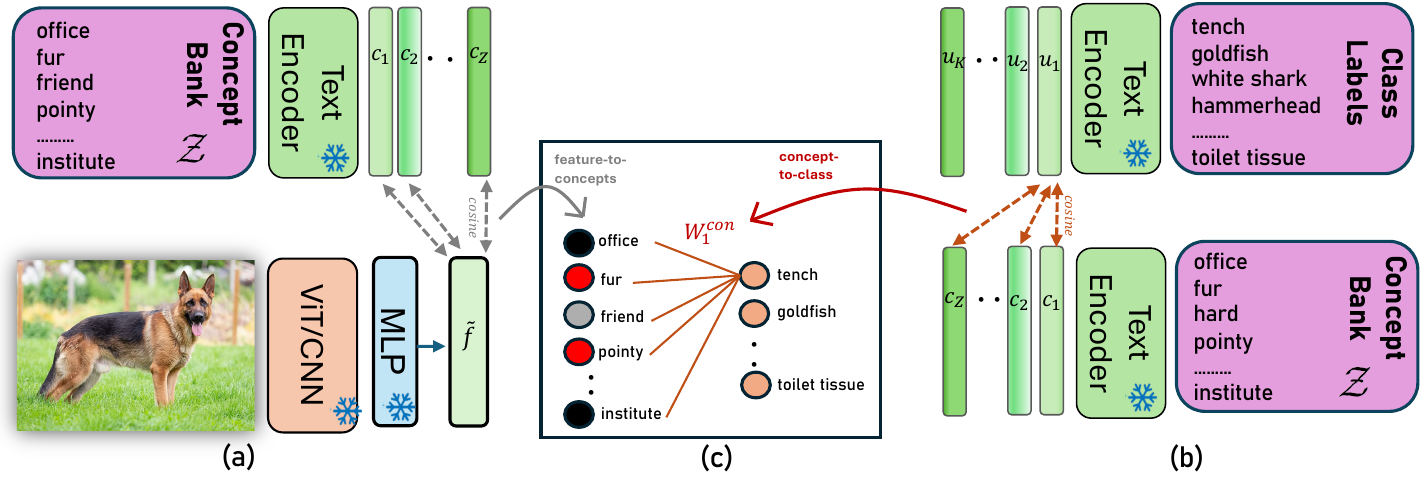}
    \caption{For any pretrained classifier, we first perform \textbf{(a)} concept discovery, followed by \textbf{(b)} building the concepts-to-class classifier in an unsupervised manner, which results in \textbf{(c)} our final U-F$^2$-CBM. Note that the concept bank only needs to be encoded once.}
    \label{zscbms}
\end{figure*}

\noindent The most straightforward way to training the \text{MLP} is to leverage the ground-truth labels from the dataset, aligning $S$ with the ground-truth distribution. However, this approach violates two key desiderata: (1) it necessitates annotated data, and (2) re-training the legacy classifier alters its original decision distribution $o$, thereby changing the reasoning process of the classifier (i.e., how it maps visual features to class probabilities and makes predictions). Notably, the original soft probability distribution $o$ is a function of the linear classifier $W$, so $W$ cannot be ignored. We instead propose to align $S$ to the original decision distribution $o$ through cross-entropy loss. For a single sample, the loss is
\begin{equation}
\label{eq:distil}
L=-\sum_{i=1}^K o_i \log \left(\frac{e^{s_i}}{\sum_{j=1}^K e^{s_j}}\right)\;.
\end{equation}
This task can be viewed as a knowledge distillation problem, except that we do not distill the knowledge of a bigger teacher model to a smaller student model, but instead \textbf{distill the distribution of the original model to its counterpart vision-language distribution}. This loss is equivalent to the KL divergence loss between $o$ and the predicted distribution, since the additional entropy term $H(o)$ that appears in the KL divergence loss is a constant that does not depend on the \text{MLP} parameters.  Eq. \ref{eq:distil} shows that our approach does not require any labels for training, and steers the \text{MLP} to preserve the classifier's distribution, since it is explicitly trained for that purpose. Note also that this loss function naturally encodes the classifier's relationship across all classes. 

\noindent We provide an illustration of the TexUnlock approach in Figure \ref{method}, and a PyTorch-like pseudocode of it in Listing \ref{lst:pseudo}. It is important to note that we only use the class name to formulate the textual prompt $l^p$, and no other supplementary information such as class descriptions, concepts or hierarchies (see Section \ref{classnames_only} in the Appendix for more information).

\vspace{1em}
\begin{lstlisting}[caption={PyTorch-like pseudocode for TextUnlock}, label={lst:pseudo}]
# text_feats: textual features of class names from a frozen sentence encoder, shape (num_classes, text_dim)
# classifier: linear classifier weights of a frozen vision_encoder, shape (visual_dim, num_classes)
# mlp: trainable MLP from visual_dim -> text_dim
# images: batch of B images, shape (N, 3, height, width)

visual_feats = vision_encoder(images)                          # (N, visual_dim)
logits       = visual_feats @ classifier                       # (N, num_classes)
original_dist = softmax(logits, dim=-1)                        # (N, num_classes)

mapped_feats = mlp(visual_feats)                            # (N, text_dim)
mapped_feats = l2_norm(mapped_feats)                           # (N, text_dim)    
text_feats = l2_norm(text_feats)                               # (N, text_dim)         

pred_logits      = mapped_feats @ text_feats.T             # (N, num_classes)
pred_dist      = softmax(pred_logits, dim=-1)              # (N, num_classes)

# cross entropy with original model's soft distribution 
loss = -(original_dist * log(pred_dist)).sum(dim=1).mean()
loss.backward()  # only mapper parameters are updated
\end{lstlisting}

\noindent After training, the projected visual features and the text encoder features lie in the same space. We can therefore query the visual features with any text by finding the alignment score between the embedded text and the projected visual features. In the case of image classification, the text queries remain the class prompts, and encoding them with the text encoder $T$ is equivalent to generating the weights of a linear classifier for the classification task formulated as \text{argmax}($\tilde{f}.U^T$), see Figure \ref{method}(b).

\subsection{U-F$^2$-CBMs}
\label{sec:zscbms}
Once the classifier’s distribution is matched to its corresponding vision–language counterpart distribution via TextUnlock, we proceed to formulate the proposed U-F$^2$-CBMs. Note that at this stage, all model components (including the \text{MLP}) are frozen and no further training is performed. A CBMs consist of two steps: (1) \textbf{concept discovery}, followed by (2) \textbf{concept-to-class prediction}. In step (1), the dense output features of a visual encoder are first mapped to textual concepts (e.g., words or short descriptions of objects) each with a score that represents the concept activation to the image. In step (2), a linear classifier $W^{con}$ is trained on top of these concept activations to predict the class. 

\noindent \textbf{Concept Discovery:} We remind readers from Section \ref{sec:method} that $U \in \mathbb{R}^{K \times m}$ is the output of the text encoder $T$ for the class prompts, which represent the classification weights of the newly formulated classifier. In CBMs, we are given a large set of textual concepts, denoted as $\mathcal{Z}$, and with cardinality $|\mathcal{Z}| = Z$. Following other works \citep{Rao2024DiscoverthenNameTC} and without loss of generality, we use the $Z=20K$ most common words in English \citep{mostcommongoogle} as our concept set. These are general concepts that are sufficiently expressive and represent world knowledge and are not tailored towards any specific dataset. We ablate on other concept sets in Appendix Section \ref{app:llm_concept_sets} for the interested reader. To ensure that the concepts are meaningful, we apply a rigorous filtering procedure to the concept set. Specifically, we remove any terms that exactly match the target class name, as well as any constituent words that form the class name (for example, eliminating “tiger” and “shark” when the class name is “tiger shark”). In addition, we exclude terms corresponding to the parent and subparent classes (e.g., “fish” and “animal” for the class “tiger shark”), other species within the same category, and any synonyms of the target class name. Details of this procedure can be found in Section \ref{app:filtering_llm_prompt} of the Appendix. This systematic filtering guarantees that the resulting concept set is free of terms that are overly similar or directly derived from the target classes. With this filtering procedure, since no concept appears in the training data, our concept discovery set becomes entirely zero-shot. This will also demonstrate the MLP’s ability to generalize to the semantic space surrounding the class name.

\noindent We use the same text encoder $T$ that generates the linear classifier $U$ to generate concept embeddings, by feeding each concept $z_i\in \mathcal{Z}$ to the text encoder $T$ to generate a concept embedding $c_i$. That is, $c_i = T(z_i)$, $\forall i=1,\dots, Z$. By performing this for all $Z$ concepts, we obtain a concept embedding matrix $C \in \mathbb{R}^{Z \times m}$. For an image $I$, we extract its visual features $f$ and use the $\text{MLP}$ to map them to $\tilde{f}$ which now lies in the text embedding space. That is, $\tilde{f} = \text{MLP}(f)$, and $\tilde{f} \in \mathbb{R}^{m}$. Since $C$ and the mapped visual features $\tilde{f}$ are now in the same space, we can query $\tilde{f}$ to find which concepts it responds to. That is, we perform concept discovery using the cosine similarity between $\tilde{f}$ and each row-vector in $C$. The concept activations are obtained by $\tilde{f}.C \in \mathbb{R}^{Z}$ and represent the activation score for each of the $Z$ concepts. Concepts that the model identifies in an image will produce a high activation score. We provide an illustration in Figure \ref{zscbms}(a). 

\begin{table}[t]
\centering
\newcommand{\cnn}{\rowcolor{cyan!10}}
\newcommand{\tr}{\rowcolor{orange!12}}
\begin{tabular}{lccc}
\toprule
\textbf{Model} & \textbf{Top-1} & \textbf{Orig.} & \textbf{$\Delta$}  \\
\midrule
\cnn ResNet50                      & 75.80 & 76.13 & $-$0.33 \\ 
\cnn ResNet50$_{\text{v2}}$        & 80.14 & 80.34 & $-$0.20  \\
\cnn ResNet101                     & 77.19 & 77.37 & $-$0.18 \\
\cnn WideResNet101$_{\text{v2}}$   & 82.21 & 82.34 & $-$0.13 \\
\cnn ResNeXt101-64x4d              & 83.13 & 83.25 & $-$0.12  \\
\cnn DenseNet161                   & 77.04 & 77.14 & $-$0.10\\
\cnn EfficientNetv2-M              & 84.95 & 85.11 & $-$0.16 \\
\cnn ConvNeXtV2-B                  & 84.56 & 84.73 & $-$0.17  \\
\cnn ConvNeXtV2-B$_{\text{pt}}$@384 & 87.34 & 87.50 & $-$0.16 \\
\midrule
\tr ViT-B/16                       & 80.70 & 81.07 & $-$0.37 \\
\tr ViT-B/16$_{\text{v2}}$        & 84.94 & 85.30 & $-$0.36 \\
\tr ViT-L/32                       & 76.72 & 76.97 & $-$0.25  \\
\tr ViT-L/16                       & 79.56 & 79.66 & $-$0.10 \\
\tr Swin-Base                      & 83.22 & 83.58 & $-$0.36 \\
\tr Swinv2-Base                    & 83.72 & 84.11 & $-$0.39 \\
\tr BeiT-L/16                      & 87.22 & 87.34 & $-$0.12 \\
\tr DINOv2-B                       & 84.40 & 84.22 & +0.18  \\
\bottomrule
\end{tabular}
\caption{\textbf{TextUnlock-ed visual classifiers for \colorbox{cyan!10}{CNNs} and \colorbox{orange!12}{Transformers} maintain classification performance.} Comparison of our re-formulated classifiers for several models (remaining models in Appendix). Top-1 indicates our results of the new formulation, and Orig. denotes the original Top-1 accuracy. $\Delta$ represents their difference ($\Delta$ = Top-1 $-$ Orig).}
\label{tab:model_performance}
\end{table}

\noindent \textbf{Concept-to-Class Prediction:} The classifier $W^{con}$ takes the concept activations produced from the concept discovery stage, and outputs a distribution $S_{cn}$ over classes. We build $W^{con}$ in an unsupervised manner; here \textit{unsupervised} indicates that no training is required to map concept activations to classes. Recall that both $U$ and $C$ are outputs of the text encoder $T$, and they are already in the same space. Therefore, we can build the weights of the classifier $W^{con}$ with a text-to-text search between the concepts and the class name. Specifically, we calculate the cosine similarity between the concept embeddings $C$ and the classification matrix $U$ to obtain the new weights for $W^{con}$. That is, we perform $C \cdot U^T \in \mathbb{R}^{Z \times K}$. Therefore, the weights of $W^{con}$ represent how similar the class name is to each of the concepts. This process is shown in Figure \ref{zscbms}(b). While the operation is performed entirely in the text feature space, the text encoder serves as the classifier weight generator. Because the MLP learns to map image representations to the text-encoder space, it is effectively mapping images into the classifier weight space. In total, the output distribution $S_{cn}$ of the CBM is obtained by feeding the identified concept activations in the image obtained from the concept discovery stage, to $W^{con}$. That is, 
\begin{equation}
\label{eq:cbms}
S_{cn} = \underbrace{(\tilde{f} \cdot C^T)}_{\text{\color{black}{concept discovery}}}\cdot  
\underbrace{(C \cdot U^T)}_{\text{\color{black}{concept-to-class}}} 
= \tilde{f} \cdot 
\underbrace{C^T C}_{\text{\color{black}{gram matrix}}}\cdot U^T\;.
\end{equation}
From Eq. \ref{eq:cbms} we make an interesting observation. Our formulation involves scaling the linear feature-based classifier $U$ by the gram matrix of concepts ($C^T C \in \mathbb{R}^{m \times m}$). The gram matrix represents a feature correlation matrix measuring how different dimensions of the feature space relate to each other. Notably, if the gram matrix is the identity ($C^T C = I$), we get back our original feature-based classifier given by $\tilde{f}.U^T$. Therefore, to convert any classifier to a CBM, we plug in the gram matrix in-between, making it a convenient way to directly switch to an inherently interpretable model. Eq. \ref{eq:cbms} also shows that we do not change the linear classifier $U$, we only scale it by the gram matrix of concepts. This means our CBMs preserve the basic reasoning process of the original classifier. By this, we obtain CBMs that discover concepts and build $W^{con}$ in an unsupervised manner for any classifier (Figure \ref{zscbms}(c)). An additional unique property is the construction of CBMs \textit{at inference time}, allowing any concept set to be selected to build a CBM on-the-fly. This makes our method highly flexible to the concept set. 

\begin{table*}[t]
\centering
\scalebox{0.99}{
\begin{tabular}{lll@{\quad}lll}
\toprule
\multicolumn{6}{c}{\textbf{Supervised CBMs}} \\
\cmidrule(lr){1-6}
\textbf{Method} & \textbf{Model} & \textbf{Top-1} & \textbf{Method} & \textbf{Model} & \textbf{Top-1} \\
\midrule
LF-CBM   & CLIP \textcolor{teal}{ResNet50}  & 67.5 & LF-CBM    & CLIP \textcolor{violet}{ViT-B/16}  & 75.4 \\
LaBo     & CLIP \textcolor{teal}{ResNet50}  & 68.9 & LaBo      & CLIP \textcolor{violet}{ViT-B/16}  & 78.9 \\
CDM      & CLIP \textcolor{teal}{ResNet50}  & 72.2 & CDM       & CLIP \textcolor{violet}{ViT-B/16}  & 79.3 \\
DCLIP    & CLIP \textcolor{teal}{ResNet50}  & 59.6 & DCLIP     & CLIP \textcolor{violet}{ViT-B/16}  & 68.0 \\
DN-CBM   & CLIP \textcolor{teal}{ResNet50}  & 72.9 & DN-CBM    & CLIP \textcolor{violet}{ViT-B/16}  & 79.5 \\
DCBM-SAM2& CLIP \textcolor{magenta}{ViT-L/14}  & 77.9 & DCBM-RCNN & CLIP \textcolor{magenta}{ViT-L/14}  & 77.8 \\
\midrule
\multicolumn{6}{c}{\textbf{Unsupervised, Label-Free, CLIP-Free CBMs (U-F$^2$-CBM)}} \\
\cmidrule(lr){1-6}
\textbf{Method} & \textbf{Model} & \textbf{Top-1} & \textbf{Method} & \textbf{Model} & \textbf{Top-1} \\
\midrule
U-F$^2$-CBM & \textcolor{teal}{ResNet50}                          & 73.9 & U-F$^2$-CBM & ViT-B/32                     & 73.3 \\
U-F$^2$-CBM & \textcolor{teal}{ResNet50$_{\text{v2}}$}          & \underline{78.1} & U-F$^2$-CBM & \textcolor{violet}{ViT-B/16}                    & 79.3 \\
U-F$^2$-CBM & ResNet101                         & 75.3 & U-F$^2$-CBM & \textcolor{violet}{ViT-B/16$_{\text{v2}}$}       & 83.2 \\
U-F$^2$-CBM & ResNet101$_{\text{v2}}$          & 79.9 & U-F$^2$-CBM & Swin-Base                    & 82.2 \\
U-F$^2$-CBM & WideResNet50                      & 76.9 & U-F$^2$-CBM & Swinv2-Base                  & 82.6 \\
U-F$^2$-CBM & WideResNet50$_{\text{v2}}$       & 79.2 & U-F$^2$-CBM & \textcolor{violet}{ViT-B/16$_{\text{pt}}$}       & \underline{81.5} \\
U-F$^2$-CBM & WideResNet101$_{\text{v2}}$      & 81.0 & U-F$^2$-CBM & BeiT-B/16                    & 83.0 \\
U-F$^2$-CBM & DenseNet121                       & 69.9 & U-F$^2$-CBM & DINOv2-B                     & 82.6 \\
U-F$^2$-CBM & DenseNet161                       & 75.2 & U-F$^2$-CBM & ConvNeXt-B$_{\text{pt}}$      & 84.0 \\
U-F$^2$-CBM & EfficientNetv2-S                  & 83.0 & U-F$^2$-CBM & ConvNeXtV2-B$_{\text{pt}}$    & 84.9 \\
U-F$^2$-CBM & EfficientNetv2-M                  & 83.9 & U-F$^2$-CBM & BeiT-L/16                    & 86.2 \\
U-F$^2$-CBM & ConvNeXt-Small                    & 81.9 & U-F$^2$-CBM & \textcolor{magenta}{ViT-L/16$_{\text{v2}}$}       & \underline{86.3} \\
U-F$^2$-CBM & ConvNeXt-Base                     & 82.8 & U-F$^2$-CBM & ConvNeXtV2-B$_{\text{pt}}$@384 & \textbf{86.4} \\
\bottomrule
\end{tabular}
}
\caption{\textbf{Our U-F$^2$-CBMs outperform CLIP-based counterparts.} Accuracy of Supervised and U-F$^2$-CBMs (ours) on ImageNet validation set. Similar backbones are color-coded. Best within backbone family is \underline{underlined}, overall best is \textbf{bold}.}
\label{tab:cbm_table_singlecol}
%\vspace{-0.5cm}
\end{table*}

\section{Experiments}
\label{sec:experiments}

We first provide results of the the classifier with its corresponding vision-language distribution aligned using our method TextUnlock. We use the most challenging CBM benchmark of ImageNet-1K dataset. There is also widespread publicly available visual classifiers trained and evaluated on ImageNet. In Section \ref{app:transformed_other_datasets} of the Appendix, we also report results on other datasets including Places365 \citep{zhou2017places}, EuroSAT \citep{helber2019eurosat} and DTD \citep{Cimpoi2013DescribingTI} showing that our method is also applicable to domain-specific, fine-grained datasets as well as datasets with a small number of classes. We apply TextUnlock on a diverse set of 40 visual classifiers. For CNNs, we consider the following family of models (each with several variants): Residual Networks (ResNets) \citep{He2015DeepRL}, Wide ResNets \citep{Zagoruyko2016WRN}, ResNeXts \citep{Xie2016AggregatedRT}, ShuffleNetv2 \citep{Ma_2018_ECCV}, EfficientNetv2 \citep{Tan2021EfficientNetV2SM}, Densely Connected Networks (DenseNets) \citep{Huang2016DenselyCC}, ConvNeXts \citep{Liu2022ACF} and ConvNeXtv2 \citep{Woo2023ConvNeXtVC}. For Transformers, we consider the following family of models (each with several variants): Vision Transformers (ViTs) \citep{dosovitskiy2021an}, DINOv2 \citep{oquab2024dinov}, BeiT \citep{bao2022beit}, the hybrid Convolution-Vision Transformer CvT \citep{Wu2021CvTIC}, Swin Transformer \citep{Liu2021SwinTH} and Swin Transformer v2 \citep{Liu2021SwinTV}. All models are pretrained on ImageNet-1K from the PyTorch \citep{torchvision2016} and HuggingFace \citep{wolf-etal-2020-transformers} libraries. Models with the subscript \textit{pt} indicate that the model was pretrained on ImageNet-21k before being finetuned on ImageNet-1K. Models with a subscript \textit{v2} are trained following the updated PyTorch training recipe \citep{PytorchNewTrainRecepie}. BEiT, DINOv2 and ConvNeXtv2 are pretrained in a self-supervised manner before being finetuned on ImageNet-1k. When training with TextUnlock, both the pretrained classifier and text encoder remain frozen, only the MLP is trained on the ImageNet training set following Eq. \ref{eq:distil}. 

\noindent Performance is evaluated using the same protocol and dataset splits as the original classifier, specifically the 50,000 validation split of ImageNet-1K. For the text encoder, we use the MiniLM Sentence Encoder \citep{wang2020minilm} as it is fast and efficient. We provide ablation studies on other text encoders in Section \ref{app:ablation_text_encoders} of the Appendix. Results for 17 classifiers are presented in Table \ref{tab:model_performance}, with the remaining 23 classifiers reported in Section \ref{app:performance_additional_models} of the Appendix. We report the replicated Top-1 accuracy of the re-formulated classifier with our method in the first column, the original Top-1 accuracy of the classifier in the second column, and the difference between them ($\Delta$) in the last column. As it can be seen, the loss in performance as indicated by $\Delta$ is minimal, with an average drop in performance of approximately $0.2$ points across all models. We also perform ablation studies on the MLP to verify its design, impact, role and that it learns meaningful transformations in Section \ref{sec:mlp_ablations} of the Appendix (see Section~\ref{subsec:mlp_role} for the role and impact of the MLP and Section~\ref{subsec:mlp_hyperparameters} for its design hyperparameters). We also evaluate our transformed classifier robustness to prompt variations in Section \ref{sec:prompt_variations} of the Appendix. For implementation details, we refer to Section~\ref{supp:implementation_details} of the Appendix.  

\noindent \textbf{U-F$^2$-CBM:} We report CBM evaluation results on the ImageNet validation set using the top-1 accuracy in Table \ref{tab:cbm_table_singlecol}. We compare against six SOTA methods: LF-CBMs \citep{oikarinen2023labelfree}, LaBo \citep{Yang2022LanguageIA}, CDM \citep{Panousis2023SparseLC}, DCLIP \citep{Menon2022VisualCV}, DN-CBMs \citep{Rao2024DiscoverthenNameTC}, and DCBM \citep{Knab2024DCBMDV}, all using the same concept set for fair comparison. All these methods are supervised and use CLIP-based models for computing the concept activations, and many use it as a backbone as well. Our U-F$^2$-CBM outperforms all the supervised CBMs, setting a new state-of-the-art performance. Notably, even a simple ResNet-50 classifier trained solely on ImageNet already outperforms the CBM for the significantly more powerful ResNet-50 CLIP model trained on 400M samples (that is, we used 400$\times$ less images). Even an EfficientNetv2-S with only 21M parameters can significantly outperform all CLIP models. It outperforms the largest CLIP ViT-L/14 model of 428M parameters by +5.1\% points, although being 20$\times$ smaller. The best results are obtained by the ConvNeXtv2 model, which achieves a top-1 accuracy of $86.4$. All models show close to original accuracy, which means we can transform any classifier to be inherently interpretable without notable performance loss. 

\paragraph{Other Datasets:} We experiment with other datasets and show that our method is applicable even to domain-specific, fine-grained datasets as well as datasets with a small number of classes. We report CBM results on Places365 (domain-specific to scenes, 365 classes), DTD (domain-specific to texture and fine-grained, 47 classes), and EuroSAT (domain-specific to satellite images, 10 classes). When a baseline method is also reported on that dataset, we include it. Otherwise, we use CLIP models as a baseline to act as a feature extractor and to compute concept activations, and train a linear classifier on top of the concept activations, formulating a CLIP supervised baseline. All baselines use the same concept set. Note also that all baselines train a supervised linear classifier, while our method derives the classifier in an unsupervised manner. Results are shown in Table \ref{tab:cbms_other_datasets_results}. For Places 365, we can see that U-F$^2$-CBM for DenseNet161 classifier (using the transformed ImageNet-only trained classifier with TextUnlock) outperforms supervised CLIP-based ResNet and ViTs CBM methods. The same applies for EuroSAT and DTD with our U-F$^2$-CBM. Therefore, the experiments show that our method scales to those scenarios as well. 

\begin{table}[h]
\centering
\begin{tabular}{lllc}
\toprule
\textbf{Dataset} & \textbf{Method} & \textbf{Model} & \textbf{Acc (\%)} \\
\midrule
\multirow{8}{*}{Places365} 
  & DCLIP     & CLIP-ResNet50   & 37.90 \\
  & DCLIP     & CLIP-ViT-B/16   & 40.30 \\
  & LF-CBM    & CLIP-ResNet50   & 49.00 \\
  & LF-CBM    & CLIP-ViT-B/16   & 50.60 \\
  & CDM       & CLIP-ResNet50   & 52.70 \\
  & CDM       & CLIP-ViT-B/16   & 52.60 \\
\cmidrule{2-4}
  & Ours      & ResNet50        & 51.57 \\
  & Ours      & DenseNet161     & \textbf{53.42} \\
\midrule
\multirow{5}{*}{EuroSAT}
  & Baseline & CLIP-ResNet50   & 86.27 \\
  & Baseline & CLIP-ViT-B/16   & 88.57 \\
\cmidrule{2-4}
    & Ours     & ViT-B/16        & 93.65 \\
    & Ours     & WideResNet101   & 94.12 \\
  & Ours     & ResNet50        & \textbf{94.22} \\
\midrule
\multirow{5}{*}{DTD}
  & Baseline & CLIP-ResNet50   & 57.77 \\
  & Baseline & CLIP-ViT-B/16   & 61.86 \\
\cmidrule{2-4}
  & Ours     & WideResNet101   & 66.97 \\
  & Ours     & ViT-B/16        & 68.46 \\
    & Ours     & ResNet50        & \textbf{68.88} \\
\bottomrule
\end{tabular}
\caption{CBM results on Places365, EuroSAT, and DTD datasets.}
\label{tab:cbms_other_datasets_results}
\vspace{-0.6cm}
\end{table}

\begin{figure*}[h]
    \centering
    \includegraphics[width=\textwidth]{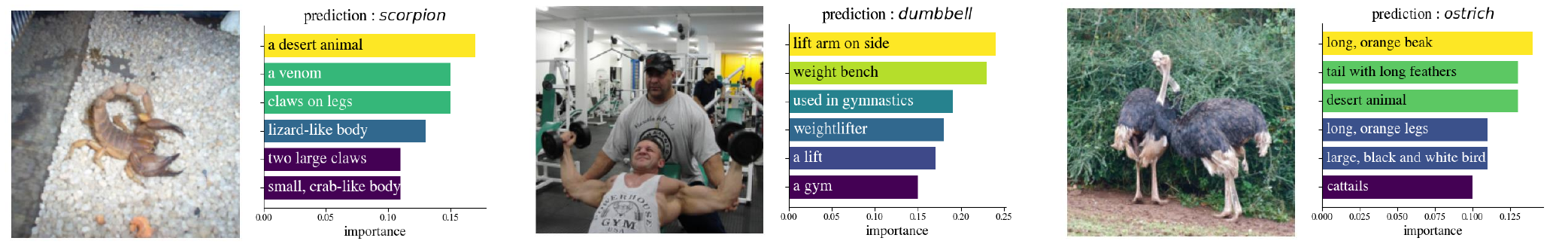}
    \caption{Qualitative examples of our U-F$^2$-CBM. We show the top-detected concepts, each with their corresponding importance score.}
    \label{cbms_qualitative}
\end{figure*}

\paragraph{Concept Interventions:} Another common way to evaluate interpretability of CBMs is concept intervention. This follows metrics from the explainability literature \cite{Petsiuk2018rise}. In this way, we can show the effectiveness of the concepts, how we can mitigate biases, debug models and fix their reasoning by explicitly intervening in the concepts of the bottleneck layer to control predictions. As these metrics serve as complementary evaluations, we refer readers to Section \ref{supp:concept_interventions} of the Appendix for the results.
\newline
\newline
\noindent In Figure \ref{cbms_qualitative}, we present qualitative examples of a selection from the top concepts responsible for the prediction, along with their weight importance on the x-axis. The weight importance is calculated by multiplying the concept activation with its corresponding weight to the predicted class. We use various concept sets to demonstrate the flexibility of our method to any desired concept set directly at test time (on-the-fly), as this process simply involves encoding the chosen concept set using the text encoder. All examples use the LF-CBM concept set \citep{oikarinen2023labelfree}. By observing the first example, the image is predicted as a ``scorpion" because it is has a lizard-like overall body, it is a venomous, desert animal with claws on its legs, and has large claws on its hand that look like a crab. Interestingly, in the second example, the top-detected concept is a “lift arm on the side.” Although this is not the primary feature defining a dumbbell, it reflects a well-documented bias in the literature \citep{Samek2019TowardsEA} regarding the ``dumbbell” class. Because most training images for this class show a dumbbell being lifted by an arm, the classifier not only learns to recognize the dumbbell but also associates it with the hand or arm that lifts it. With our method, we can obtain a textual interpretation of the biases that the original classifier learns. We also provide qualitative examples of global class-wise concepts detected in Section \ref{app:global_classwise_example} of the appendix.

\section{Zero-Shot Image Captioning}
We also show that TextUnlock enables zero-shot image captioning with any pretrained visual classifier beyond CLIP models. Existing zero-shot captioning methods largely rely on CLIP and its shared vision–language space. Thanks to TextUnlock, zero-shot image captioning can now be performed for any pretrained visual classifier. We adapt the method introduced in ZeroCap \citep{Tewel2021ZeroCapZI} for our purpose. Specifically, ZeroCap is a test-time approach that learns to produce a caption that maximizes the similarity with the image features. We first project the visual feature vector $f$ using the \text{MLP} to obtain $\tilde{f}$. That is, $\tilde{f} = \text{MLP}(f)$. Since $\tilde{f}$ is now in the same space as the text encoder $T$, we can measure its association to any encoded text. We utilize an off-the-shelf pretrained language decoder model, denoted as $G$, to generate open-ended text. We keep $G$ frozen to maintain its language generation capabilities and instead use prefix-tuning \citep{Li2021PrefixTuningOC} to guide $G$ to generate a text that maximizes the similarity with the transformed visual feature vector $\tilde{f}$. More details about ZeroCap is in Section \ref{supp:process_decoding_feats} of the appendix.

\noindent We now evaluate the performance of the zero-shot captions produced on the COCO image captioning dataset \citep{Lin2014MicrosoftCC}. Since we do not train any model on the ground-truth image captions provided by COCO, we use zero-shot image captioning as a benchmark. Note that the COCO dataset differs in distribution than ImageNet, as a single image may contain many objects, interactions between them, and categories not included in ImageNet (\textit{e.g., }person). Therefore, it also serves as a way to evaluate generalization of our method to other datasets, given that we only used the ImageNet images and class names for training. We present results on the widely used ``Karpathy test split" with various vision classifiers. As baselines, we compare our approach against existing methods in zero-shot image captioning, specifically ZeroCap \citep{Tewel2021ZeroCapZI} and ConZIC \citep{Zeng2023ConZICCZ}, both which use CLIP. For evaluation, we employ standard natural language generation metrics: BLEU-4 (B@4) \citep{Papineni2002BleuAM}, METEOR (M) \citep{Banerjee2005METEORAA}, ROUGE-L (R-L) \citep{lin-2004-rouge}, CIDEr (C) \citep{Vedantam2014CIDErCI}, and SPICE (S) \citep{Anderson2016SPICESP}. Results are shown in Table~\ref{tab:captioning_scores}. ConvNeXtv2 achieves state-of-the-art performance on CIDEr and SPICE, the two most critical metrics for evaluating image captioning systems. Even with a simple ResNet-50 vision encoder trained on ImageNet-1K (1.2 million images), our approach outperforms the baseline methods on CIDEr and SPICE, despite the latter utilizing the significantly more powerful CLIP vision encoder, trained on 400 million image-text pairs (that is, we used 400$\times$ less images compared to CLIP). Qualitative examples of the produced zero-shot image captions from different visual classifiers are shown in Section \ref{suppsec:captioning_qualitative} of the Appendix. Note that our results in Table \ref{tab:captioning_scores} are outperformed by the baseline ZeroCap on the BLEU-4 (B4) and METEOR (M) metrics. However, it is important to note that B4 and M are n-gram overlap-based metrics. They assume that the generated caption follows a specific structure and style. We verify this hypothesis by applying compositional image captioning \citep{Kulkarni2011BabyTU, Lu2018NeuralBT} with in-context learning.  With this technique, the last row of Table \ref{tab:captioning_scores} with the superscript \textit{com} shows that the (B4, M and R-L) are boosted, which verifies our hypothesis about the low scores of B4 and M compared to baseline methods. We refer to Section \ref{app:compositional_captioning} of the Appendix for more details.

\begin{table}
  \centering
  \scalebox{0.9}{
    \begin{tabular}{lccccc}
      \toprule
      \textbf{Model} & \textbf{B4} & \textbf{M} & \textbf{R-L} & \textbf{C} & \textbf{S} \\
      \midrule
      ZeroCap              & \textbf{2.6}   & \textbf{11.5} & \textemdash & 14.6 & 5.5 \\
      ConZIC               & 1.3   & 11.5 & \textemdash & 12.8 & 5.2 \\
      \midrule
      \multicolumn{6}{c}{\textbf{Ours}} \\
      \midrule
      %DenseNet161          & 1.50 &  10.2  & 20.4 & 15.8  &  6.3  \\
      ResNet50             & 1.43 &  10.2  & 20.3 & 15.9  &  6.2  \\
      ResNet50$_{\text{v2}}$           & 1.47 &  10.5  & 20.6 & 16.8  &  6.5  \\
      %WideResNet50         & 1.40 &  10.2  & 20.4 & 16.0  &  6.4  \\
      %WideResNet101$_{\text{v2}}$     & 1.50 &  10.4  & 20.5 & 16.6  &  6.4  \\
      %ResNet101$_{\text{v2}}$          & 1.48 &  10.4  & 20.6 & 16.7  &  6.5  \\
      ConvNeXt-B$_{\text{pt}}$      & 1.50 &  10.6  & 20.8 & 17.2  &  6.7  \\
      DINOv2-Base          & 1.50 &  10.7  & 21.0 & 17.3  &  6.7  \\
      %ViT-B/16$_{\text{v2}}$           & 1.50 &  10.5  & 20.9 & 17.3  &  6.5  \\
      BeiT-L/16            & 1.50 &  10.6  & 20.9 & 17.6  & 6.9 \\
      ViT-B/16$_{\text{pt}}$        & 1.50 &  10.7  & 20.9 & 17.7  & 6.9 \\
      ConvNeXtV2-B$_{\text{pt}}$@384& 1.60 & 10.7  & \textbf{21.1} & \textbf{17.9}  &  \textbf{6.9}  \\ \hline
      ConvNeXtV2-B$_{\text{pt}}$@384$^{\text{com}}$ & \textbf{4.40}   & \textbf{12.7}   & \textbf{30.2}  & \textbf{18.7}   & \textbf{7.2} \\
      \bottomrule
    \end{tabular}
  }
  \caption{Zero-Shot Image Captioning Performance}
  \label{tab:captioning_scores}
\end{table}

\section{Conclusion}
We introduced a method for transforming any frozen visual classification model into a CBM. We proposed TextUnlock, the core of our method, that aligns the distribution of the original classifier with that of its vision–language counterpart. This allows us to then produce an unsupervised, label-free and CLIP-free (U-F$^2$) CBM, which outperforms supervised CLIP-based CBMs across 40 models. We also showed zero-shot image captioning as an additional application. Finally, as with any research work, this study has its own limitations, discussed in Section \ref{appendix:limitations} of the Appendix.

{
    \small
    \bibliographystyle{ieeenat_fullname}
    \bibliography{main}
}

\clearpage
\setcounter{page}{1}
\maketitlesupplementary

\setcounter{section}{0}
\setcounter{figure}{0}
\setcounter{table}{0}

\section{Limitations}
\label{appendix:limitations}

As with any research work, our study has its own limitations that should be transparent and acknowledged. In particular, we identify a primary limitation of our method concerning wrong semantic associations of class names in CBMs, due to polysemy. This issue is closely tied to the choice of concept set used. When using the 20K most common words in English as our concept set, we observe that class names get associated to wrong semantically-related concepts. For example, the bird ``drake" is linked to artist-related concepts (Rihanna, Robbie, lyric). This occurs because the bird ``drake" is less familiar to the text encoder than the artist ``drake". In fact, a google search with the word “drake” yields directly the artist rather than the bird. As another example, the top-detected concepts for the prediction “african grey” are incorrect semantic associations with the word “african” (ethiopian, tanzania) and the word "grey" (purple, blue) that do not pertain to the bird itself. We also observe similar cases that may raise ethical concerns (e.g., the animal ``cock" leads to associations with male reproductive terms). However, it is worth noting the following: 

\begin{enumerate}
 \item Meaningful concepts such as ``duck" (first example) are still detected among the top concepts.
 \item The incorrect semantic associations contribute only a negligible portion of the total logit, accounting for approximately 0.01\% of the overall prediction score.
 \item This issue is considerably less severe when using alternative concept sets, such as the LF-CBM concept set tailored for ImageNet.
\item This issue also appears in CLIP-based CBMs and hence not unique to our approach.
\end{enumerate}

\section{Ablation Studies on the Text Encoder}
\label{app:ablation_text_encoders}

In Table \ref{tab:text_encoder_performance}, we present ablation studies using other text encoders from the Sentence-BERT library \citep{reimers-2019-sentence-bert} with the ResNet50 visual classifier. We observe that the choice of the text encoder has minimal effect on the performance. This is because even lower-performing text encoders are capable of understanding class names. 

\begin{table}[h]
    \centering
    \begin{tabular}{lc}
        \toprule
        \textbf{Text Encoder} & \textbf{Top-1 (\%)} \\
        \midrule
        DistilRoberta         & 75.73 \\
        MPNet-Base            & 75.78 \\
        MPNet-Base-MultiQA   & 75.76 \\
        MiniLM           & \textbf{75.80} \\
        \bottomrule
    \end{tabular}
    \caption{Ablation studies on other text encoders}
    \label{tab:text_encoder_performance}
\end{table}

\section{Ablation Studies of the MLP}
\label{sec:mlp_ablations}

\subsection{Role and Impact of the MLP}
\label{subsec:mlp_role}
We first evaluate the role and impact of the MLP to verify its contribution and that it learns meaningful transformations, and does not collapse into a trivial function. We perform the following ablation studies: \textbf{1) Mean Ablation:} For an image, we replace the input features to the MLP with a constant mean of the features calculated across the full ImageNet validation set. \textbf{2) Random Features:} For an image, we replace the input features to the MLP with random values sampled from a normal distribution with a mean and standard deviation equal to that of the features calculated across the full ImageNet validation set. \textbf{3) Random Weight Ablation:} We randomize the weights of the MLP projection. \textbf{4) Shuffled Ablation:} For an image, we replace the input features to the MLP with input features of another random image in the validation dataset. 

\noindent For each ablation experiment, we compute the ImageNet validation accuracy. We expect the accuracy to drop across all ablations. As shown in Table \ref{tab:mlp_ablations}, the accuracy nearly drops to zero in every case, confirming the effectiveness of our MLP.

\begin{table*}
    \centering
    \scalebox{0.99}{
    \begin{tabular}{lccccc}
        \toprule
        Model &  & Mean Feature $\downarrow$ & Random Features $\downarrow$ & Shuffled Features $\downarrow$ & Random Weights $\downarrow$ \\
        \midrule
        \multirow{2}{*}{ResNet101v2} 
            & Ours    & 81.49 & 81.49 & 81.49 & 81.49 \\
            & Ablated & \textbf{0.10}  & \textbf{0.11}  & \textbf{1.70}  & \textbf{0.11}  \\ \hline
        \multirow{2}{*}{ConvNeXt-Base} 
            & Ours    & 83.88 & 83.88 & 83.88 & 83.88 \\
            & Ablated & \textbf{0.10}  & \textbf{0.11}  & \textbf{1.79}  & \textbf{0.10}  \\ \hline
        \multirow{2}{*}{BeiT-L/16} 
            & Ours    & 87.22 & 87.22 & 87.22 & 87.22 \\
            & Ablated & \textbf{0.10}  & \textbf{0.11}  & \textbf{1.87}  & \textbf{0.11}  \\ \hline
        \multirow{2}{*}{DINOv2-B} 
            & Ours    & 84.40 & 84.40 & 84.40 & 84.40 \\
            & Ablated & \textbf{0.10}  & \textbf{0.13}  & \textbf{1.76}  & \textbf{0.09}  \\
        \bottomrule
    \end{tabular}
    } 
    \caption{Ablation studies of the MLP.}
    \label{tab:mlp_ablations}
\end{table*}

\subsection{MLP Design}
\label{subsec:mlp_hyperparameters}

Next, we perform an ablation study over the number of layers and output dimension scaling factor (\texttt{dim\_out\_factor}) of the MLP. \texttt{dim\_out\_factor} specifies how much to scale the previous layer’s dimensionality. For example, in a ViT-B/16 model the hidden dimension is 768; setting \texttt{dim\_out\_factor = 2} therefore expands the projection from 768 to 2 $\times$ 768 = 1536. Results are shown in Table \ref{tab:mlp_ablations_hyper} for a ResNet50 model. We observe that the 2 layers and a output dimension scaling factor of 2 provides the best results.

\begin{table}[h]
\centering
\begin{tabular}{ccc}
\toprule
\textbf{Layers} & \textbf{\texttt{dim\_out\_factor}} & \textbf{Top-1 (\%)} \\
\midrule
1 & 1 & 72.48 \\
1 & 2 & 74.01 \\
2 & 1 & 75.41 \\
2 & 2 & 75.80 \\
\bottomrule
\end{tabular}
\caption{Performance comparison of the MLP for different layer and dimension configurations.}
\label{tab:mlp_ablations_hyper}
\end{table}

\section{Concept Interventions and other Explainability Metrics} 
\label{supp:concept_interventions}

\subsection{Concept Interventions}
We report concept intervention results on our ZS-CBMs. Interventions on CBMs are an effective tool to mitigate biases, debug models and fix their reasoning by explicitly intervening in the concepts of the bottleneck layer to control predictions. The  Waterbirds-100 dataset \citep{Sagawa2020Distributionally} is a standard dataset used in previous works \citep{Rao2024DiscoverthenNameTC} to conduct CBM intervention experiments. It is a binary classification dataset of two classes: waterbirds and landbirds. The training images of waterbirds are on water backgrounds, and training images of landbirds are on land backgrounds. However, the validation images do not have that correlation, where waterbird images appear on land backgrounds and landbird images on water backgrounds. The model is assumed to learn the water–land background correlation to perform this classification task. By building a CBM, we can correct this bias by intervening in concepts in the CB layer. We curated a validation dataset of waterbirds/landbirds directly from the ImageNet validation set, including 140 validation images (70 for each class). We create our ZS-CBM using the two class prompts:``an image of a waterbird" (for the waterbird class) and ``an image of a landbird" (for the landbird class), and using the same concept set from \citep{Rao2024DiscoverthenNameTC} which includes a collection of bird-related concepts and a collection of land-related concepts. The ZS-CBM achieves a low accuracy, as shown in Table \ref{tab:intervention}, which indicates the water-land bias the model performs for classification. To correct this, we intervene in the concepts in the CB layer, following the setup from \citep{Rao2024DiscoverthenNameTC}. \textbf{Intervention R (Int. R):} We zero-out activations of any bird concepts from the bottleneck layer, and expect the accuracy to drop. The larger the drop, the better. \textbf{Intervention K (Int. K):} We keep activations of bird concepts as they are, but scale down the activations of all remaining concepts by multiplying them with a factor of 0.1, and expect the accuracy to increase. More increase is better. Results are presented in Table \ref{tab:intervention} for some models, and demonstrate the success of our intervention experiments. 

\noindent We also conduct concept intervention experiments with a more challenging multi-class setup. Unlike the Waterbirds dataset where bias‑correlation issues are assumed, we make no such assumption here. We instead evaluate whether  intervening on class-related concepts in the bottleneck layer and zeroing out their activations, impairs model accuracy. A drop in accuracy indicates that these concepts are important for prediction. We select a subset of 10 classes from ImageNet following \citep{Howard_Imagenette_2019}. We use the original ImageNet validation images for those classes, rather than the validation set from \citep{Howard_Imagenette_2019}, because the latter contains original ImageNet training examples that our model has already seen. We employ an LLM to generate 5 highly-relevant concepts for each class, achieving in total 50 concepts. The classes and concepts we used are provided in Section \ref{supp:cbm_multiclass_classes_concepts} of the supplementary material. We then measure CBM accuracy before and after intervention. Results are provided in Table \ref{tab:mc-intervention}. For each image, we zero out the activations of its class-related concepts and report the \textbf{Intervention R (Int. R)} metric. As shown in Table \ref{tab:mc-intervention}, this intervention reduces accuracy by approximately 20\% on average, underscoring the concepts importance to the CBM.

\begin{table}[h]
  \centering
  \scalebox{0.8}{
      \begin{tabular}{lccc}
        \toprule
        Model & Orig. CBM & Int.\,(R)$\downarrow$ & Int.\,(K)$\uparrow$ \\
        \midrule
        BeiT-B/16                        & 54.29 & 41.43 \textbf{(-12.86)} & 58.57 \textbf{(+4.28)} \\
        ConvNeXtV2$_{\text{pt}}$@384     & 53.57 & 42.14 \textbf{(-11.43)} & 59.29 \textbf{(+5.72)} \\
        ConvNeXt\_B$_{\text{pt}}$        & 53.57 & 42.86 \textbf{(-10.71)} & 58.57 \textbf{(+5.00)} \\
        DiNOv2                           & 52.86 & 43.57 \textbf{(-9.29)}  & 59.29 \textbf{(+6.43)} \\
        BeiT-L/16                        & 52.86 & 44.29 \textbf{(-8.57)}  & 58.57 \textbf{(+5.71)} \\
        \bottomrule
      \end{tabular}
      }
    \caption{CBM Interventions on the standard Waterbirds dataset}
    \label{tab:intervention}
  \end{table}

\begin{table}[h]
  \centering
      \begin{tabular}{@{}lcc@{}}
        \toprule
        Model & Orig.\ CBM & Int.\,(R)$\downarrow$ \\
        \midrule
        ResNet50                          & 96.80 & 76.00 \textbf{(-20.8)}  \\
        ResNet101                         & 97.00 & 76.80 \textbf{(-20.2)}  \\
        DINOv2-B                          & 98.60 & 78.40 \textbf{(-20.2)}  \\
        BeiT-B/16                         & 98.60 & 78.80 \textbf{(-19.8)}  \\
        ConvNeXtV2$_{\text{pt}}$@384      & 99.40 & 79.40 \textbf{(-20.0)}  \\
        \bottomrule
      \end{tabular}
    \caption{CBM Interventions in a multi-class setup}
    \label{tab:mc-intervention}
\end{table}

\subsection{Concept Discovery Accuracy}
We use the discrete-based Mutual Information (MI) metric\footnote{https://github.com/fawazsammani/clip-interpret-mutual-knowledge} to calculate the MI between concepts detected by our method, and those detected by CLIP. Specifically, given the concept bank, we use the CLIP Vision Encoder to encode an image, and the CLIP Text Encoder to encode the concept bank. We record the set of top-K detected concepts most similar to the image. We then encode the same image with our TextUnlock-ed visual classifier (e.g., DINOv2)(including the \text{MLP}) and encode the same concept bank with our Text Encoder $T$. We record the set of top-K detected concepts most similar to the image. We then calculate the MI between the detected concepts from CLIP (CLIP ViT-B/16 model) and our method. Results are shown in Table \ref{tab:nmi_across_models} and demonstrate high mutual information which indicates that our method successfully detects key concepts in the dataset by only learning from the limited vocabulary of class labels. 

\begin{table}[h]
    \centering
    \begin{tabular}{l c}
        \toprule
        \textbf{Model}        & \textbf{NMI} \\
        \midrule
        DINOv2                & 0.681 \\
        EfficientNetv2-M      & 0.704 \\
        ViT-B/16v2            & 0.711 \\
        \bottomrule
    \end{tabular}
    \caption{Normalized Mutual Information (NMI) with CLIP ViT-B/16 for different models. NMI ranges from 0 (no mutual information) to 1 (perfect agreement).}
    \label{tab:nmi_across_models}
\end{table}

\section{Transformed Classifier Results on Other Datasets}
\label{app:transformed_other_datasets}
We conduct extra experiments on 3 datasets. Places365 (domain-specific to scenes), DTD (domain-specific to texture and fine-grained), and EuroSAT (domain-specific to satellite images). For each dataset, we show the top-1 accuracy of the classifier's original performance and the top-1 accuracy of our transformed classifier. Results are shown in Table \ref{tab:other_datasets_acc1_transfomred}. 

\begin{table}[h]
\centering
\scalebox{0.85}{
\begin{tabular}{llcc}
\toprule
\textbf{Dataset} & \textbf{Model} & \textbf{Original (\%)} & \textbf{Transformed (\%)} \\
\midrule
Places365 & ResNet50      & 54.77 & 53.90 \\
          & DenseNet161   & 56.13 & 55.54 \\
\midrule
EuroSAT   & ResNet50      & 93.95 & 94.23 \\
          & WideResNet101 & 93.78 & 93.88 \\
          & ViT-B/16      & 93.67 & 93.53 \\
\midrule
DTD       & ResNet50      & 69.47 & 69.26 \\
          & WideResNet101 & 68.62 & 68.14 \\
          & ViT-B/16      & 69.95 & 69.68 \\
\bottomrule
\end{tabular}
}
\caption{Results of our transformed classifier on Places365, EuroSAT, and DTD datasets}
\label{tab:other_datasets_acc1_transfomred}
\end{table}

\section{Prompt Variations}
\label{sec:prompt_variations}

We evaluate the robustness of our transformed classifier to variations in text prompts. Using the transformed ViT-B/16, we provide a diverse set of prompts and measure the Top-1 accuracy on ImageNet. Results are shown in  Table \ref{tab:prompt_variation}. We order the prompts from highest to lowest scoring. We find that our transformed classifier is robust to text prompts.   The worst prompt degrades the baseline accuracy by only 0.36 points

\begin{table}[h]
\centering
\begin{tabular}{lc}
\toprule
\textbf{Prompt} & \textbf{Top-1 Accuracy (\%)} \\
\midrule
an image of a \{\}                & 80.70 \\
a photo of one \{\}.              & 80.67 \\
a photo of a \{\}.                & 80.66 \\
a close-up photo of a \{\}.       & 80.63 \\
a black and white photo of a \{\} & 80.63 \\
a close-up photo of the \{\}.     & 80.62 \\
a cropped photo of a \{\}.        & 80.61 \\
a good photo of a \{\}.           & 80.61 \\
a dark photo of a \{\}.           & 80.61 \\
a bright photo of a \{\}.         & 80.60 \\
a bright photo of the \{\}.       & 80.59 \\
a bad photo of a \{\}.            & 80.57 \\
a blurry photo of a \{\}.         & 80.57 \\
a pixelated photo of a \{\}.      & 80.55 \\
a photo of many \{\}.             & 80.54 \\
a low-resolution photo of the \{\}& 80.54 \\
a low-resolution photo of a \{\}. & 80.52 \\
a photo of my \{\}.               & 80.45 \\
a jpeg-corrupted photo of a \{\}. & 80.34 \\
\bottomrule
\end{tabular}
\caption{Effect of different text prompts on Top-1 accuracy using ViT-B/16.}
\label{tab:prompt_variation}
\end{table}

\section{Other Concept Sets}
\label{app:llm_concept_sets}

Early works such as label-free CBMs use LLMs to automatically generate a set of concepts for a target class. In recent works such as DN-CBM \citep{Rao2024DiscoverthenNameTC} and DCBM \citep{Knab2024DCBMDV}, it was however shown that general, pre-defined concept sets outperform the LLM-generated ones, which introduce many spurious correlations and hallucinated associations. In Table \ref{tab:llm_vs_20k_concepts}, we compare general pre-defined and LLM-generated concept sets. Across different types of architectures, general, pre-defined concept sets outperform the LLM-generated ones by several percent. Please note that while earlier works such as LF-CBMs employ LLM-generated concept sets (which underperform compared to the general, pre-defined concept set), all results reported in our main manuscript—including those from prior works—are presented using the general, pre-defined concept set to ensure a fair comparison. 

\begin{table}[h]
\centering
\begin{tabular}{lcc}
\toprule
\textbf{Model} & \textbf{Predefined} & \textbf{LLM-generated} \\
\midrule
Swinv2-Base       & 82.60 & 80.34 \\
ConvNext-Base     & 82.80 & 80.24 \\
ConvNeXtV2-B$_{\text{pt}}$@384 & 86.40 & 84.03 \\
ViT-B/16          & 79.30 & 76.37 \\
\bottomrule
\end{tabular}
\caption{Comparison of Predefined and LLM-generated concept sets on CBMs}
\label{tab:llm_vs_20k_concepts}
\end{table}

\section{Implementation Details}
\label{supp:implementation_details}
For the text encoder, we use the all-MiniLM-L12-v1\footnote{https://huggingface.co/sentence-transformers/all-MiniLM-L12-v1} model available on the Sentence Transformers library \citep{reimers-2019-sentence-bert}. This text encoder was trained on a large and diverse dataset of over 1 billion training text pairs. It contains a dimensionality of $m=384$ and has a maximum sequence length of 256. 

\noindent Our \text{MLP} projector is composed of 3 layers, the first projects the visual feature dimensions $n$ to $n \times 2$ and is followed by a Layer Normalization \citep{Ba2016LayerN}, a GELU activation function \citep{Hendrycks2016GaussianEL} and Dropout \citep{Srivastava2014DropoutAS} with a drop probability of 0.5. The second layer projects the $n \times 2$ dimensions to $n \times 2$ and is followed by a Layer Normalization and a GELU activation function. The final linear layer projects the $n \times 2$ dimensions to $m$ (the dimensions of the text encoder). We train the \text{MLP} projector with a batch size of 256 using the ADAM optimizer \citep{KingBa15} with a learning rate of 1e-4 that decays using a cosine schedule \citep{loshchilov2017sgdr} over the total number of epochs. We follow the original image sizes that the classifier was trained on. 

\noindent For the training images, we apply the standard image transformations that all classifiers were trained on which include a Random Resized Crop and a Random Horizontal Flip. For the validation images, we follow exactly the transformations that the classifier was evaluated on, which include resizing the image followed by a Center Crop to the image size that the classifier expects. Each model is trained on a single NVIDIA GeForce RTX 2080 Ti GPU.

\section{Compositional Image Captioning}
\label{app:compositional_captioning}

Compositional Image Captioning is an old image captioning paradigm termed \citep{Kulkarni2011BabyTU}, later revived with deep learning methods \citep{Lu2018NeuralBT}. In compositional captioning, a set of image-grounded concepts (such as attributes, objects and verbs) are first detected, and a language model is then used to compose them into a natural sounding sentence. With the current advancements of Large Language Models (LLMs) and their powerful capabilities, we use an LLM as a compositioner. Specifically, we detect the top concepts and verbs to the image using the concept discovery method introduced in Section \ref{sec:zscbms} and shown in Figure~\ref{zscbms}(a), and feed them, along with their similarity scores, to an LLM. For the concepts, we use the same concept set as in Section \ref{sec:zscbms}. We also add a list of the most common verbs \citep{commonverbs} in English to the pool. This allows us to cover all possible words and interactions. We prompt the LLM to utilize the provided information to compose a sentence, given a limited set of in-context examples from the COCO captioning training set (in our experiments, we use 6 examples). This allows us to generate sentences adhering to a specific style and structure of the in-context examples. For this experiment, we used GPT4o-mini \citep{gpt4omini} as our LLM, as it is fast and cost-efficient. In the prompt, we explicitly instruct the LLM to refrain from reasoning or generating content based on its own knowledge or assumptions, and that all its outputs must be strictly grounded in the provided concepts, verbs, and score importance. We use the following prompt:

\texttt{I will give you several attributes and verbs that are included in an image, each with a score. The score reflects how important (or how grounded) the attrribute/verb is to the image, and higher means more important and grounded. Your job is to formulate a caption that describes the images by looking at the attributes/verbs with their associated scores. You should not reason or generate anything that is based on your own knowledge or guess. Everything you say has to be grounded in the attrbiutes/verbs and score importances. Please use the following the structure, style, and pattern of the following examples. Example 1: A woman wearing a net on her head cutting a cake. Example 2: A child holding a flowered umbrella and petting a yak. Example 3: A young boy standing in front of a computer keyboard. Example 4: a boy wearing headphones using one computer in a long row of computers. Example 5: A kitchen with a stove, microwave and refrigerator. Example 6: A chef carrying a large pan inside of a kitchen. Here are the attributes and scores: \{detected concepts with scores\}, and these are the verbs and scores: \{detected verbs with scores\}.}

Results are shown in Table \ref{tab:compositional_captions_20k}. 

\begin{table}[!ht]
  \centering
  \scalebox{0.9}{
    \begin{tabular}{lccccc}
      \toprule
      \textbf{Model} & \textbf{B4} & \textbf{M} & \textbf{R-L} & \textbf{C} & \textbf{S} \\
      \midrule
      ZeroCap               & 2.6           & 11.5          & \textemdash & 14.6          & 5.5          \\
      ConZIC                & 1.3           & 11.5          & \textemdash & 12.8          & 5.2          \\
      \midrule
      \multicolumn{6}{c}{\textbf{Ours}} \\
      \midrule
      DenseNet161           & 4.20          & 12.5          & 30.1          & 17.0          & 6.6          \\
      ResNet50              & 4.10          & 12.5          & 30.1          & 17.0          & 6.6          \\
      ResNet50$_{\text{v2}}$ & 4.50          & 12.8          & 30.5          & 18.4          & 6.9          \\
      WideResNet50$_{\text{v2}}$        & 4.20          & 12.7          & 30.3          & 17.7          & 6.9          \\
      ResNet101v2           & 4.30          & 12.6          & 30.1          & 18.0          & 6.8          \\
      ConvNeXt-Base$_{\text{v2}}$   & 4.40     & 12.8          & 30.2          & 18.6          & 7.1          \\
      ResNet50$_{\text{v2}}$            & 4.50          & 12.8          & 30.5          & 18.4          & 6.9          \\
      EfficientNetv2-S      & 4.40          & 12.7          & 30.4          & 18.6          & 6.9          \\
      ViT-B/16$_{\text{pt}}$        & 4.50          & 12.8          & 30.2          & 18.7          & \textbf{7.2} \\
      ConvNeXtV2-B$_{\text{pt}}$@384 & 4.40          & 12.7          & 30.2          & 18.7          & \textbf{7.2} \\
      BeiT-B/16             & 4.50          & 12.8          & 30.3          & \textbf{18.9} & 7.1          \\
      DINOv2-Base           & \textbf{4.60} & \textbf{13.0} & \textbf{30.7} & 18.7          & 7.1          \\
      \bottomrule
    \end{tabular}
  }
  \caption{Zero-Shot Compositional Captioning Performance}
  \label{tab:compositional_captions_20k}
\end{table}

\noindent While the results on CiDEr and SPICE are incremental compared to the results in Table \ref{tab:captioning_scores}, the n-gram metrics (B4, M and R-L) are boosted, which verifies our hypothesis that the low scores of B4 and M were attributed to the specific caption style and structure of the respective dataset (COCO). 

\subsection{Domain-Specific Compositional Captioning}
\label{subsec:domain_captioning}
Since the LLM we used for compositional image captioning is generic and can adapt to any style and structure, compositional captioning is especially useful for generating captions tailored to specific domains. We explore alternative concept sets in Compositional Captioning. In the main manuscript, we reported results using the 20,000 most common English words as our concept set. Since the LLM remains fixed and functions as a composer, integrating detected concepts and verbs grounded in the image into a caption, we can seamlessly substitute the concept set with any domain-specific concept set alternative. This allows for the generation of captions tailored to a specific domain. Here, we maintain the same set of verbs but explore the use of concepts specific to the ImageNet dataset. Since ImageNet lacks dedicated captions, we evaluate the domain-specific captioning by anticipating a decline in performance on the COCO captioning dataset. We use the ImageNet-specific concept set from \citep{oikarinen2023labelfree} and report zero-shot captioning performance in Table \ref{tab:compositional_captions_lfset}. As shown, we observe a decrease in all metrics. This shows that our method can readily produce captions for any domain. Finally, also note that we can control the style of the generations by simply prompting the LLM to compose the concepts and verbs in a specific style (e.g., humorous, positive, negative).

\begin{table}
\centering
\begin{tabular}{lccccc}
\toprule
\textbf{Method} & \textbf{B4} & \textbf{M} & \textbf{R-L} & \textbf{C} & \textbf{S} \\
\midrule
ZeroCap          & 2.6      & 11.5     & \textemdash & \textbf{14.6} & 5.5     \\
ConZIC           & 1.3      & 11.5     & \textemdash & 12.8     & 5.2     \\
\midrule
\multicolumn{6}{c}{\textbf{Ours}} \\
\midrule
MobileNetv3-L    & 3.50     & 12.7     & 29.1     & 11.4     & 6.1     \\
ResNet50         & 3.60     & 12.7     & 29.3     & 12.0     & 6.0     \\
ResNet101v2      & 3.50     & 12.9     & 29.3     & 12.2     & 6.3     \\
WideResNet101v2  & 3.70     & 12.9     & 29.6     & 12.4     & 6.2     \\
ConvNeXt-Base    & 3.80     & 12.8     & 29.5     & 12.7     & 6.2     \\
EfficientNetv2-S & 3.70     & 12.9     & 29.6     & 12.9     & 6.3     \\
ViT-B/16 (pt)    & 3.80     & 13.1     & 29.5     & 13.2     & 6.5     \\
BeiT-L/16        & \textbf{3.90} & \textbf{13.2} & \textbf{29.6} & 13.4     & \textbf{6.6} \\
\bottomrule
\end{tabular}
\caption{Composition Captioning Performance using the ImageNet-specific LF-CBM concept set}
\label{tab:compositional_captions_lfset}
\end{table}

\section{Qualitative examples of zero-shot image captioning.}
\label{suppsec:captioning_qualitative}
\begin{figure*}
    \centering
    \includegraphics[width=\textwidth]{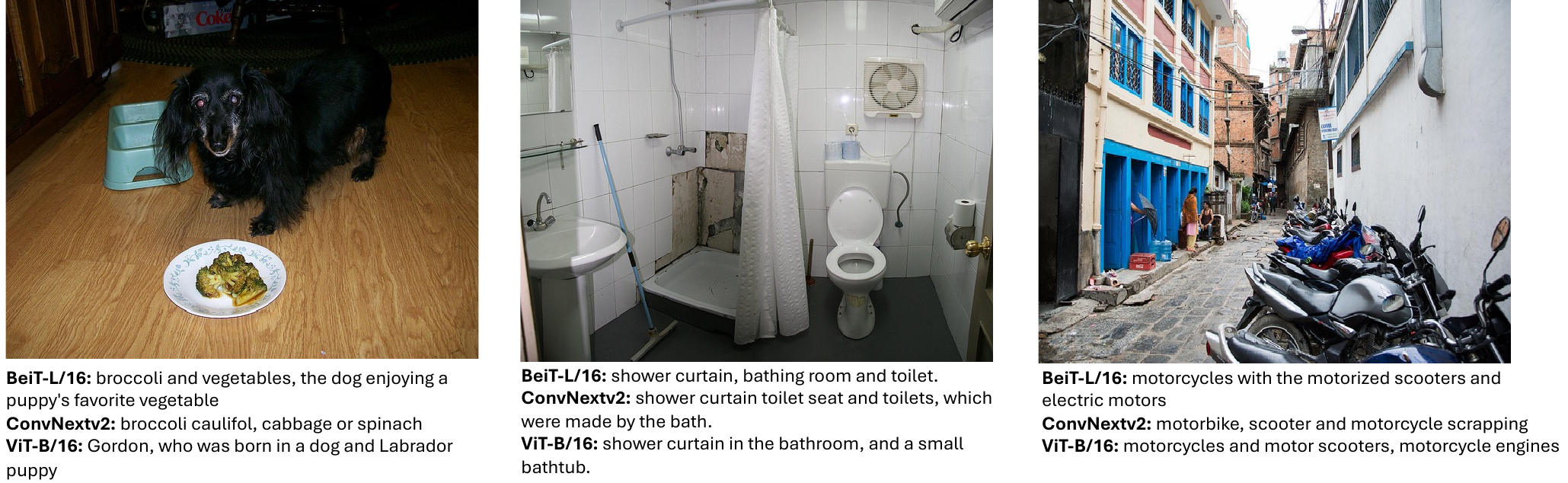}
    \caption{Qualitative examples of zero-shot image captioning.}
    \label{captioning_qualitative}
\end{figure*}

We present qualitative examples of zero-shot image captioning from different classifiers in Figure \ref{captioning_qualitative}. We choose BeiT-L/16, ConvNeXtv2, and ViT-B/16. This allows us to see how different classifiers ``see" the image. From the first example, BeiT-L/16 captures features of both the vegetables and the dog, whereas ConvNeXtV2 captures only the vegetables. In contrast, ViT-B/16 focuses exclusively on the dog and its characteristics. 

% \section{Human Evaluation of Image Captioning}
% \label{app:human_eval_captioning}
% We conducted a human-user study to evaluate the zero-shot captions produced. We considered three metrics: \textbf{1) Grounding}, which measures if the sentence and concepts it contains are visually grounded in the image, \textbf{2) Relevancy}, measures the correctness of the sentence and its relevancy to the image , and \textbf{3) Friendliness}, measures how human-friendly it is for a non-expert to understand and use this sentence to understand a model. Note that the later penalizes grammatical errors, as this hinder the users’ understanding of the sentence. We choose a random set of 50 samples using three models. Results are shown in Table \ref{tab:human_eval_captioning}. Note that the human evaluation aligns well with the metrics we used in Table \ref{tab:side_by_side_captioning} in the main manuscript. 

% \begin{table}[h]
% \centering
% \begin{tabular}{lccc}
% \toprule
% \textbf{Model} & \textbf{Grounding} & \textbf{Relevancy} & \textbf{Friendliness} \\
% \midrule
% ResNet50    & 69.83 & 62.71 & 79.90 \\
% BeiT-L      & 71.42 & 65.34 & 80.80 \\
% ConvNextv2  & 73.10 & 68.02 & 81.30 \\
% \bottomrule
% \end{tabular}
% \caption{Human Evaluation of Models on Grounding, Relevancy, and Friendliness}
% \label{tab:human_eval_captioning}
% \end{table}

\section{Performance on Additional Models}
\label{app:performance_additional_models}

We report performance on additional models that were not included in the main manuscript in Table \ref{app:more_model_performance}. 

\begin{table}
\centering
\begin{tabular}{lccc}
\toprule
\textbf{Model} & \textbf{Top-1} & \textbf{Orig.} & \textbf{$\Delta$} \\
\midrule
ConvNeXtV2-B$_{\text{pt}}$     & 86.07 & 86.25 & $-$0.18 \\
BeiT-B/16                      & 84.54 & 85.06 & $-$0.52  \\
WideResnet50                  & 78.35 & 78.47 & $-$0.12 \\
ViT-L/16$_{\text{v2}}$        & 87.61 & 88.06 & $-$0.45  \\
EfficientNetv2-S              & 84.04 & 84.23 & $-$0.19 \\
ConvNeXt-Small                & 83.42 & 83.62 & $-$0.20\\
ResNeXt101-32x8d               & 79.10 & 79.31 & $-$0.21 \\
ShuffleNetv2$_{\text{x2.0}}$  & 75.83 & 76.23 & $-$0.40 \\
WideResNet50$_{\text{v2}}$    & 81.17 & 81.31 & $-$0.14 \\ 
DenseNet169                   & 75.46 & 75.60 & $-$0.14 \\
ConvNeXt-Tiny                 & 82.19     & 82.52     & $-$0.33 \\
ConvNeXt-B$_{\text{pt}}$      & 85.27 & 85.52 & $-$0.25 \\ 
ResNet101$_{\text{v2}}$       & 81.50 & 81.68 & $-$0.18 \\ 
ResNeXt50-32x4d               & 77.44 & 77.62 & $-$0.18 \\
ConvNeXt-Base                 & 83.88 & 84.06 & $-$0.18 \\
ResNeXt50-32x4d$_{\mathrm{v2}}$ & 80.79 & 80.88 & $-$0.09 \\
ViT-B/32                      & 75.40     & 75.91     & $-$0.51 \\
Swin-Small                    & 82.63     & 83.20     & $-$0.57 \\
Swinv2-Tiny                   & 81.44     & 82.07     & $-$0.63 \\
CvT-21                        & 80.45     & 81.27     & $-$0.82 \\
Swinv2-Small                  & 83.32     & 83.71     & $-$0.39 \\
%ViT-B/16$_{\text{pt}}$        & 83.55     & 84.37     & $-$0.82 \\ 
\bottomrule
\end{tabular}
\caption{Performance of our reformulated classifiers for additional models}
\label{app:more_model_performance}
\end{table}

\section{Process of Zero-Shot Image Captioning}
\label{supp:process_decoding_feats}

We remind readers of the mapping function, denoted as $\text{MLP}$, that transforms the visual features $f$ into the same space as textual features, producing $\tilde{f}$. A pre-trained language model $G$ is then optimized to generate a sentence that closely aligns with $\tilde{f}$. To preserve the generative power of $G$, we keep it frozen and apply prefix-tuning \citep{Li2021PrefixTuningOC}, which prepends learnable tokens in the embedding space. We follow a test-time training approach to optimize learnable tokens for each test input on-the-fly. Our method builds upon the work of \citep{Tewel2021ZeroCapZI}.

\begin{figure*}[h]
    \centering
    \includegraphics[width=\textwidth]{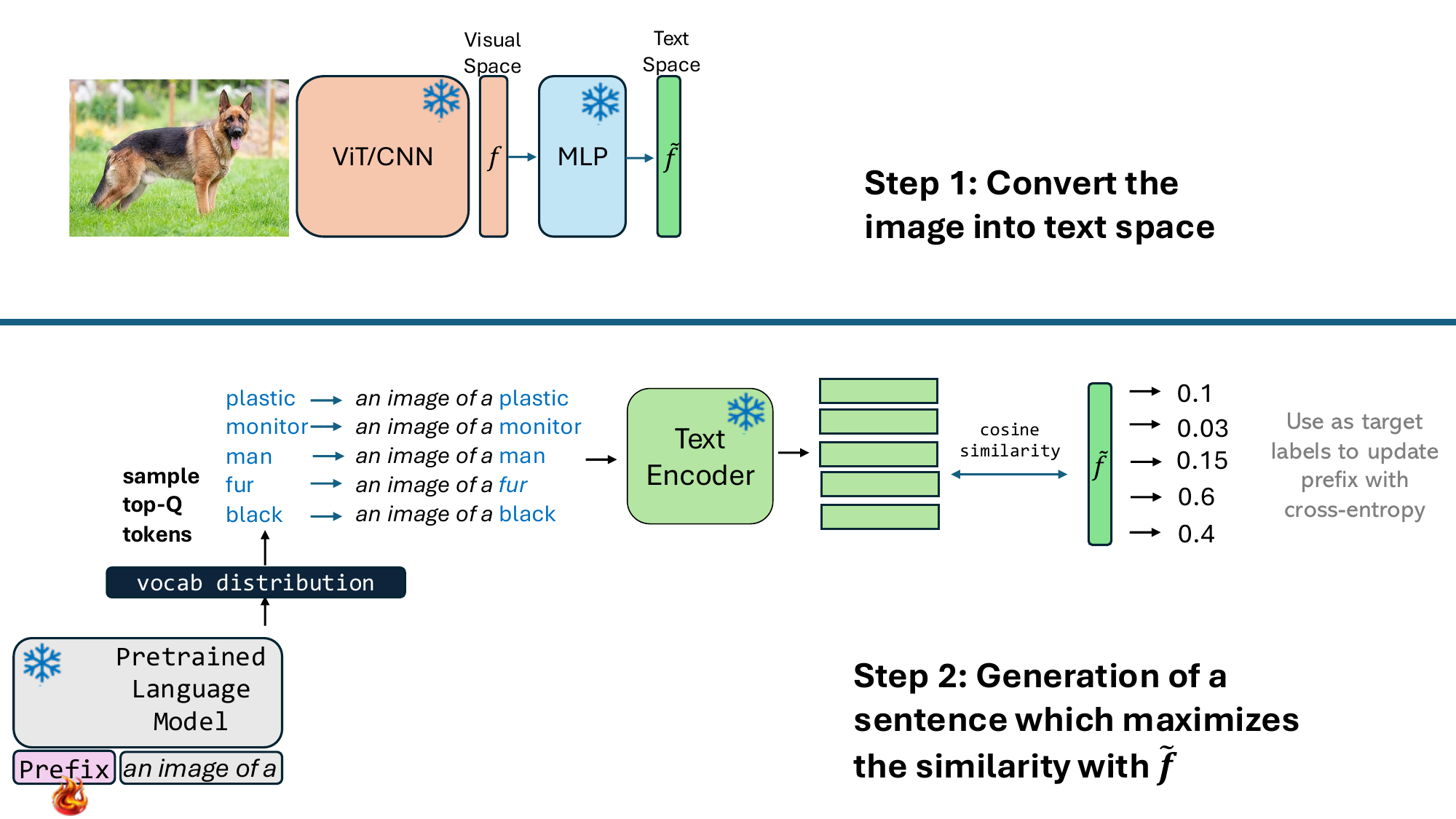}
    \caption{The process used to generate zero-shot captions using any pretrained language decoder (e.g., GPT-2). The process is shown for the first timestep ($ts=1$) and first iteration ($j=1$) with a hard prompt set as ``an image of a". We apply prefix tuning while keeping the language decoder frozen, generating text that maximizes the similarity with the visual features.} 
    \label{decoding_method_supp}
\end{figure*}

\noindent A high-level overview of this process is illustrated in Figure \ref{decoding_method_supp}. Using a pre-trained language model $G$, we prepend randomly initialized learnable tokens, referred to as prefixes, which guide $G$ to produce text that maximizes alignment with visual features. These learnable prefixes function as key-value pairs in each attention block, ensuring that every generated word can attend to them. 

\noindent For each iteration $j$, at a timestep $ts$, we sample the top-$Q$ tokens from the output distribution of $G$, denoted as $G_{out}$, which serve as possible continuations for the sentence. These $Q$ candidate sentences are then encoded by a text encoder $T$, mapping them into the same embedding space as $\tilde{f}$. We compute the cosine similarity between each encoded sentence and $\tilde{f}$, resulting in $Q$ similarity scores. These scores are normalized with softmax and define a target distribution used to train $G_{out}$ via Cross-Entropy loss. The learnable prefixes are updated through backpropagation. 

\noindent With the updated prefixes, $G$ is run again, and the most probable token is selected as the next word. This process is repeated for a predefined number of timesteps (up to the desired sentence length) or until the $<.>$ token is generated. At the end of each iteration, a full sentence is generated. We conduct this process for 20 iterations, generating 20 sentences in total. The final output is chosen as the sentence with the highest similarity to the visual features $\tilde{f}$.

\noindent We also add the fluency loss from \citep{Tewel2021ZeroCapZI} as well as other token processing operations. We refer readers to \citep{Tewel2021ZeroCapZI} for more information. We use the smallest GPT-2 of 124M parameters as $G$. We also noticed that using a bigger $G$ (e.g., GPT-2 medium) does not enhance performance, indicating that a decoder with basic language generation knowledge is sufficient.

\section{Concept Filtering}
\label{app:filtering_llm_prompt}
One of our concept filtering procedures requires us to know the terms corresponding to the parent and subparent classes (e.g., “fish” and “animal” for the class “tiger shark”), other species within the same
category, and any synonyms of the target class name. In order to obtain this information, we used an LLM (gpt-4o-mini) with the following prompt:

\noindent \texttt{Provide your answer to below as single-word comma separated. If you need to provide a term composed of 2 words, then separate each into a single word. For each class, provide its synonyms and closely related names (e.g., other species of the category, its superclasses such as bird, fish, dog, cat, animal...etc). Here is an example. The ImageNet class is: tench, tinca tinca. Answer: fish,animal,cyprinid,carp,vertebrate.}

\section{Using only class names}
\label{classnames_only}
We remind readers from Section \ref{sec:method} that we only use the class names to formate the text prompt for the text encoder when training the \text{MLP}. In practice, we can go beyond class names by using resources like a class hierarchy from WordNet \citep{Lin1998WordNetAE} (the original source where ImageNet was extracted from), or class descriptions extracted from a LLM as in CuPL \citep{Pratt2022WhatDA} or VCVD\citep{Menon2022VisualCV}. However, this approach would be considered as ``cheating. The original classifier implicitly learns the semantics, hierarchies, relationships and distinctive features of different classes. Explicitly providing additional information would not replicate the classifier faithfully since it would force the classifier to focus on predefined features or those we intend it to learn. Moreover, this would also leak information to downstream tasks such as CBMs and textual decoding of visual features, compromising the fairness of evaluation. For instance, if class descriptions were used in the training, the concepts in CBMs would align with those specified in the training prompts. For these reasons, we refrain from using any other additional information than the class names. We use the class names provided from \url{https://gist.github.com/yrevar/942d3a0ac09ec9e5eb3a}. 

\section{Global Class-Wise Qualitative Examples}
\label{app:global_classwise_example}

\begin{figure*}
    \centering
    \includegraphics[width=\textwidth]{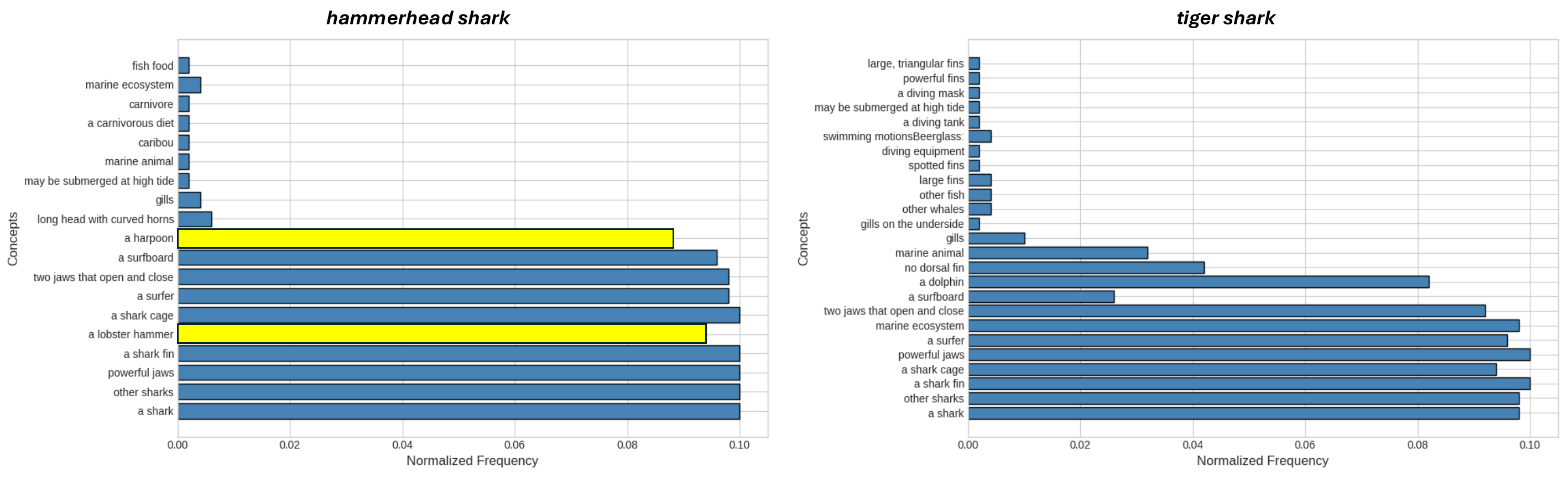}
    \caption{Global class-wise interpretability analysis with our Concept Bottleneck Model. We highlight in yellow the top concepts in "hammerhead shark" that are not present in "tiger shark", and therefore distinctive to "hammerhead shark".}
    \label{global_cbms}
\end{figure*}

In Figure \ref{global_cbms}, we present a probability distribution of global class-wise concepts. These are concepts detected for all images of a specific class, along with their frequency. We consider two semantically similar classes but distinctively different: ``hammerhead shark" and a ``tiger shark". We highlight in yellow the top concepts in ``hammerhead shark" that are not present in ``tiger shark". These concepts are ``harpoon" and ``lobster hammer", both which are distinctive to the head of the hammerhead shark and drive its prediction.  Note that for this experiment, we do not apply the concept filtering procedure. 

\section{Multi-class CBM Intervention Classes and Concepts}
\label{supp:cbm_multiclass_classes_concepts}

\begin{table*}
  \centering
  \begin{tabular}{@{}lp{10cm}@{}}
    \toprule
    \textbf{Class} & \textbf{Concepts} \\
    \midrule
    tench & fish, freshwater, fins, dorsal, olive \\
    english springer & dog, long ears, brown and white, playful, hunting \\
    cassette player & portable, audio, tape, speakers, buttons \\
    chainsaw & sharp, handheld, cutting, metal, wood \\
    church & cross, tower, architecture, sacred, religious \\
    french horn & curved, mouthpiece, musical instrument, orchestral, blow \\
    garbage truck & large vehicle, wheels, clean, high load, lift \\
    gas pump & fueling, hose, metallic, gasoline, handle \\
    golf ball & small, white, round, rubber, dimples \\
    parachute & fabric, fly, air, landing, strings \\
    \bottomrule
  \end{tabular}
  \caption{ImageNet classes and their five associated concepts we use in our multi-class CBM intervention experiment.}
  \label{tab:imagenet_classes_and_concepts_multiclass_cbm}
\end{table*}

\end{document}